\pdfoutput=1

\documentclass[11pt]{article}

\usepackage[final]{acl}

\usepackage{times}
\usepackage{latexsym}

\usepackage[T1]{fontenc}

\usepackage[utf8]{inputenc}

\usepackage{microtype}

\usepackage{inconsolata}

\usepackage{graphicx}

\usepackage{adjustbox}
\usepackage{algorithm}
\usepackage{algorithmicx}
\usepackage[noend]{algpseudocode}
\usepackage{amsfonts}
\usepackage{amsmath}
\usepackage{amsthm}
\usepackage{arydshln}
\usepackage{bbm}
\usepackage{booktabs}
\usepackage{cite}
\usepackage{subfigure}
\usepackage{url}
\usepackage{comment}
\usepackage{multirow}

\usepackage{tablefootnote}
\usepackage{etoolbox}

\usepackage{listings} 
\usepackage{booktabs}

\usepackage{bm}

\newcommand{\textred}[1]{\textcolor{red}{#1}}

\usepackage{chngcntr}
\counterwithout{footnote}{section} 

\usepackage{float} 

\title{Hallucinated Span Detection with \\ Multi-View Attention Features}

\author{Yuya Ogasa\thanks{Currently with LY Corporation, Japan. Email: yogasa@lycorp.co.jp} \\
  Grad. Sch. of Information Science and Tech. \\
  The University of Osaka\\
  Japan \\
  \texttt{ogasa.yuya@ist.osaka-u.ac.jp}
  \And
  Yuki Arase \\
  School of Computing\\
  Institute of Science Tokyo \\
  Japan \\
  \texttt{arase@c.titech.ac.jp} \\
  }

\begin{document}
\maketitle
\begin{abstract}
This study addresses the problem of hallucinated span detection in the outputs of large language models. 
It has received less attention than output-level hallucination detection despite its practical importance. 
Prior work has shown that attentions often exhibit irregular patterns when hallucinations occur. 
Motivated by these findings, we extract features from the attention matrix that provide complementary views capturing (a)~whether certain tokens are influential or ignored, (b)~whether attention is biased toward specific subsets, and 
(c)~whether a token is generated referring to a narrow or broad context, in the generation.  
These features are input to a Transformer-based classifier to conduct sequential labelling to identify hallucinated spans. 
Experimental results indicate that the proposed method outperforms strong baselines on hallucinated span detection with longer input contexts, such as data-to-text and summarisation tasks. 
\end{abstract}

\section{Introduction}
Large Language Models (LLMs) have significantly advanced natural language processing and demonstrated high performance across tasks~\citep{minaee2024largelanguagemodelssurvey}. 
However, hallucinations persisting in texts generated by LLMs have been identified as a serious issue, which undermines LLM safety~\citep{Ji_2023}.  

To tackle this challenge, hallucination detection has been actively studied~\citep{Huang_2024}. 
Model-level (e.g., \citep{min-etal-2023-factscore}) or response-level (e.g., \citep{manakul-etal-2023-selfcheckgpt}) hallucination detection has been proposed. 
However, identification of the hallucinated span is less explored despite its practical importance. 
Hallucinated span detection enables understanding and manually editing the problematic portion of the output. 
It also provides clues to mitigate hallucinations in LLM development.

To address this, we tackle hallucinated span detection. 
While there have been various types of hallucinations~\citep{wang-etal-2024-answer}, this study targets hallucinations on contextualised generations that add baseless and contradictive information against the given input context. 
Motivated by the findings that irregular attention patterns are observed when hallucination occurs~\citep{chuang-etal-2024-lookback,zaranis-etal-2024-analyzing}, we extract features to characterise the distributions of attention weights. 
Specifically, the proposed method extracts an attention matrix from an LLM by inputting a set of prompt, context, and LLM output of concern. 
It then assembles features for each token from the attention matrix: average and diversity of incoming attention as well as diversity of outgoing attention, which complementarily capture the attention patterns of language models. 
The former two features indicate whether attention is distributed in a balanced manner for tokens in the output text. 
The last feature reveals if an output token was generated by broadly attending to other tokens. 
These features are then fed to a Transformer encoder with a conditional random field layer on top to conduct sequential labelling to determine whether a token is hallucinated or not.

Experimental results on hallucinated span detection confirmed that the proposed method outperforms strong baselines on data-to-text and summarisation tasks, improving token-level F$1$ score for $4.9$ and $2.9$ points, respectively.  
An in-depth analysis reveals that the proposed method is capable of handling longer input contexts. 
Our code is available at \url{https://github.com/Ogamon958/mva_hal_det}.

\begin{figure*}[t]
    \centering
    \includegraphics[scale=0.12]
    {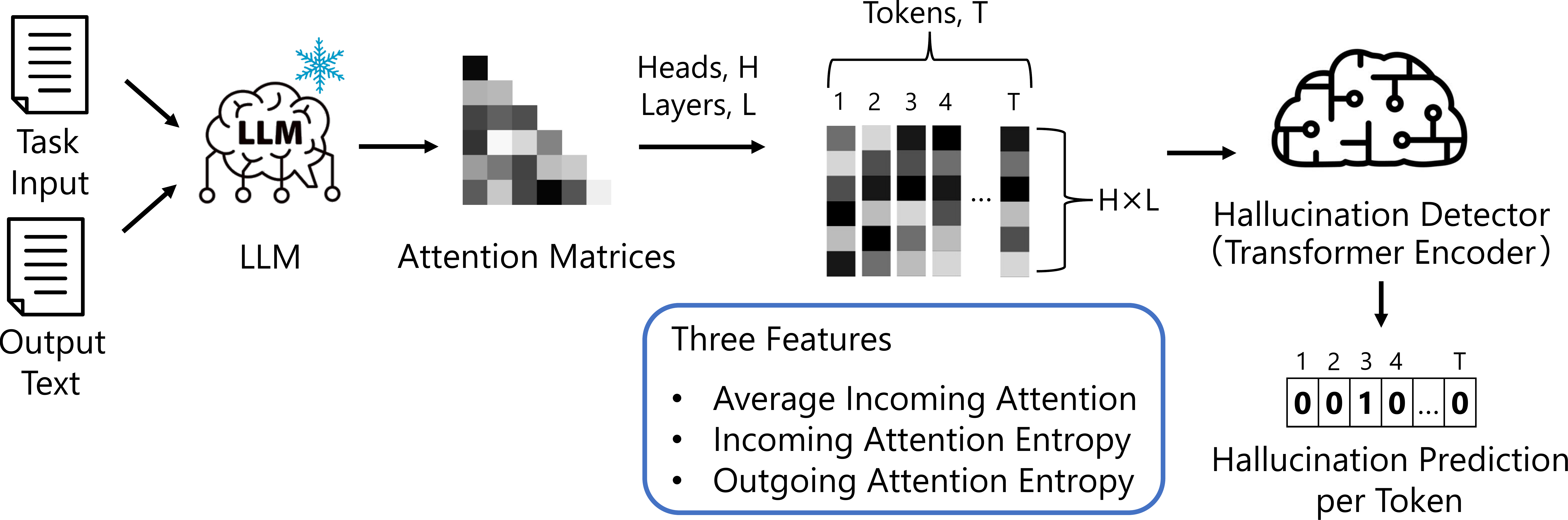}
    \caption{Overview of the proposed method}
    \label{fig:proposed_method}
\end{figure*}

\section{Related Work}
\label{sec:related_research}

This section discusses hallucination detection that utilises various internal states of LLMs. 

\paragraph{Attention-Based Hallucination Detection}
Lookback Lens~\citep{chuang-etal-2024-lookback} is the most relevant method to our study, which identifies hallucinations using only attention matrices. 
It computes the ``Lookback'' ratio of attention to assess whether generated tokens attend well to the input context. 
In contrast, our features primarily focus on the attention of output texts and capture more nuanced and structural attention patterns. 
ALTI+~\citep{ferrando-etal-2022-towards,zaranis-etal-2024-analyzing} tracks token interactions across layers. 
ALTI+ has been applied to hallucination detection in machine translation, highlighting cases where the model fails to properly utilise source text information.
A drawback of ALTI+ is its computational cost. 
It computes a token-to-token contribution matrix for each layer and for each attention head. 
Therefore, memory consumption linearly increases depending on the length of context and output as well as LLM sizes. 
Indeed, \citet{zaranis-etal-2024-analyzing} excluded sequences longer than $400$ tokens due to GPU memory constraints.

\paragraph{Other Internal States for Hallucination Detection}
Hallucination detection has also explored various internal states of LLMs other than attention.  
\citet{xiao-wang-2021-hallucination} and \citet{zhang-etal-2023-enhancing-uncertainty} identify hallucinations as tokens generated with anomalously low confidence based on the probability distribution in the final layer.  
\citet{azaria-mitchell-2023-internal} and \citet{ji-etal-2024-llm} use layer-wise Transformer block outputs to estimate hallucination risk. 
These studies assume that hallucination detection will be conducted on the same LLM generating output and can access such Transformer block outputs. 
In contrast, we empirically showed that the proposed method can also be applied to closed LLMs.
Further, attention-based methods are distinctive from these studies in that they aim to model inter-token interactions. 

\begin{figure*}[t]
    \centering
    \includegraphics[scale=0.088]
    {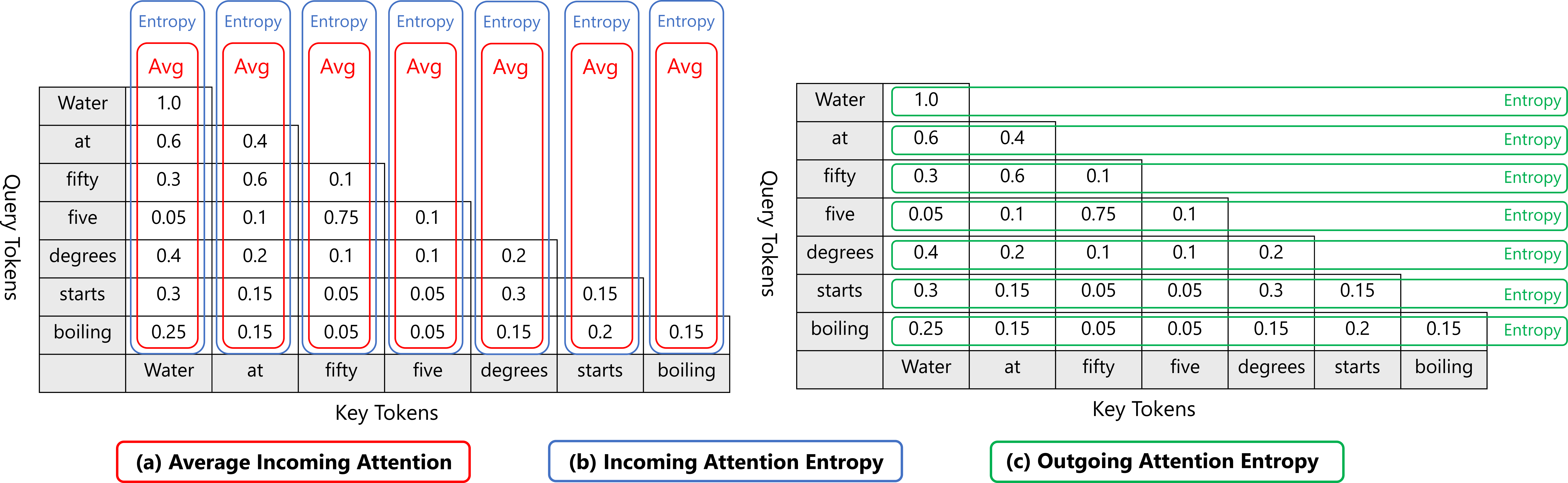}
    \caption{Feature extraction from attention matrix (these attention values are for illustrative purposes.)}
    \label{fig:proposed_features}
\end{figure*}


\section{Proposed Method}
The proposed method is illustrated in Figure \ref{fig:proposed_method}. 
It conducts sequential labelling, i.e., predicts binary labels that indicate whether a token in text, which has been generated by a certain LLM, is hallucinated or not. 
Specifically, the proposed method takes a set of prompt, input context, and output generated by an LLM of concern as input to another LLM and obtains the attention matrix of the output text span. 
It then extracts features from the attention matrix (Sections~\ref{sec:conversion_attention} and~\ref{subsubsec:feature_construction}). 
These features are fed to a Transformer encoder model with the prediction head of a conditional random field (CRF) to conduct sequential labelling to identify hallucinated spans (Section \ref{training_detector}). 
As the attention matrix provides crucial information for our method, we compare the raw attention and a variation based on the analysis of attention mechanism~\citep{kobayashi-etal-2020-attention} (Section \ref{sec:adaptation_kobayashi}). 
We remark that only the hallucination detection model needs training, i.e., the LLM for attention matrix extraction is kept frozen, which makes our method computationally efficient.

Our method applies to both scenarios where the LLM that generated outputs and the LLM for hallucinated span detection are the same or different. 
In practice, the latter setting is expected to be more common in an era where LLMs are widely used for writing tasks. 
In addition, we cannot access the internal state of proprietary LLMs. 
Our experiments assume the scenario where the LLM for generation and the LLM for detection are different.

\subsection{Feature Design}
\label{sec:conversion_attention}
Previous studies revealed that irregular patterns of attention are incurred when hallucination occurs~\citep{chuang-etal-2024-lookback,zaranis-etal-2024-analyzing}. 
Based on these findings, we design features to complementarily capture irregular attentions. 
Specifically, we extract features providing complementary views of the attention matrix as shown in Figure~\ref{fig:proposed_features}: (a) average attention a token receives (\textbf{Average Incoming Attention}), (b) diversity of attention a token receives (\textbf{Incoming Attention Entropy}), and (c) diversity of tokens that a token attends to (\textbf{Outgoing Attention Entropy}).

\paragraph{Average Incoming Attention}
We compute the average attention weights that a token receives when generating others. 
This feature indicates whether certain tokens are influential or ignored in generation. 
Specifically, it computes the average attention weight in the key direction on the attention matrix as illustrated on the left side of Figure~\ref{fig:proposed_features}.   

\paragraph{Incoming Attention Entropy}
This feature captures the diversity of attention weights, i.e., whether attention is biased toward specific subsets or is more uniformly distributed. 
It computes the entropy of attention weights in the key direction on the attention matrix as illustrated on the left side of Figure~\ref{fig:proposed_features}.  

\paragraph{Outgoing Attention Entropy}
The final feature models the diversity of tokens that a token attends to when being generated. 
This indicates whether the model references a narrow or broad range of context for generating the token. 
Specifically, this feature computes the entropy of attention weights in the query direction on the attention matrix as illustrated on the right side of Figure~\ref{fig:proposed_features}.   

Given the complex and diverse nature of attention dynamics, we do not regard individual features as independently effective. 
Rather, we assume these features \emph{complementary} capture irregular attention patterns due to hallucination by providing views from different angles. 

\subsection{Feature Extraction}
\label{subsubsec:feature_construction}
We extract these features for each token from the attention matrix. 
As notation, the output by an LLM to detect hallucinated span consists of $T$ tokens. 
The LLM for attention matrix extraction consists of $L$ layers of a Transformer decoder with $H$ heads of multi-head attention. 

\paragraph{Average Incoming Attention}
This feature computes the average attention weights that a token receives when generating other tokens.
The attention matrix \(\bm{A}\) is lower triangular due to masked self-attention, meaning each query token \(i\) attends only to key tokens \(j\) with \(1 \leq j \leq i\). 
Thus, earlier tokens receive attention more often, and tokens close to the end receive attention less often. 
To compensate for the imbalanced frequency, we adjust the attention weights $\alpha_{i,j}$ as: 
\begin{equation}
\label{eq:alpha_scale}
\alpha'_{ij} = \alpha_{ij} \cdot i.
\end{equation}
Using the adjusted attention matrix \(\bm{A}'\), the average attention that a key token \(j\) receives is computed as:
\begin{equation}
\label{eq:avg_key}
\mu_j^{(\ell,h)} = \frac{1}{T - j + 1} \sum_{i=j}^{T} \alpha_{ij}'^{(\ell,h)},
\end{equation}
where $1 \leq \ell \leq L$ is the layer index and $1 \leq h \leq H$ is the head index. 
The final feature vector is obtained by concatenating the average attention weights across all layers and heads:
\begin{equation}
\label{eq:feature_vec}
\bm{v}(j) = \bigl[\mu_j^{(1,1)},\mu_j^{(1,2)},\ \dots,\mu_j^{(L,H)}\bigr] \in \mathbb{R}^{LH}
\end{equation}

\paragraph{Incoming Attention Entropy}
To model the diversity of attention a token receives, we use the entropy of the weights. 
As discussed in the previous paragraph, the attention matrix is lower triangular. 
To compensate for different numbers of times to receive attention, we normalise an entropy value by dividing by the maximum entropy:

\begin{align} 
\beta_ j^{(\ell,h)} &= \frac{- \sum_{i=j}^{T} \kappa_{ij}^{(\ell,h)} \log \kappa_{ij}^{(\ell,h)}}{\log(T - j + 1)}, \label{eq:E_key} \\
\kappa_{ij}^{(\ell,h)} &= \frac{\alpha_{ij}'^{(\ell,h)}}{\sum_{k=1}^{i} \alpha_{ik}'^{(\ell,h)}}. \label{eq:kappa_def}
\end{align}
The final feature vector is a concatenation of the entropy values across layers and heads:
\begin{equation}
\label{eq:E_key_vec}
\bm{e}(j) = \bigl[ \beta_j^{(1,1)},\ \beta_j^{(1,2)},\ \dots,\ \beta_j^{(L,H)} \bigr] \in \mathbb{R}^{LH}
\end{equation}

\paragraph{Outgoing Attention Entropy}
This feature models the diversity of tokens that a token attends to when being generated. 
Similar to the ``Incoming Attention Entropy'' feature, we compute the entropy of attention weights of query tokens\footnote{Remind that attention weights are normalised in the query direction.} by dividing by the maximum entropy:
\begin{equation}
\label{eq:E_query}
\gamma_i^{(\ell,h)} = \frac{- \sum_{j=1}^{i} \alpha_{ij}^{(\ell,h)} \log \alpha_{ij}^{(\ell,h)}}{\log(i)}.
\end{equation}
The final feature vector is a concatenation of the entropy values across layers and heads: 
\begin{equation}
\label{eq:E_query_vec}
\bm{\hat{e}}(i) = \bigl[ \gamma_i^{(1,1)},\ \gamma_i^{(1,2)},\ \dots,\ \gamma_i^{(L,H)} \bigr] \in \mathbb{R}^{LH}
\end{equation}

\paragraph{Final Feature Vector}
The three features \(\bm{v}(j)\) (Average Incoming Attention), \(\bm{e}(j)\) (Incoming Attention Entropy), and \(\bm{\hat{e}}(i)\) (Outgoing Attention Entropy) are concatenated as a final feature vector for hallucination detection.
Each feature has \(LH\) elements; thus, the final feature vector consists of \(3LH\) elements.

\subsection{Hallucination Detector}
\label{training_detector}
\begin{figure}[t]
    \centering
    \includegraphics[scale=0.11]
    {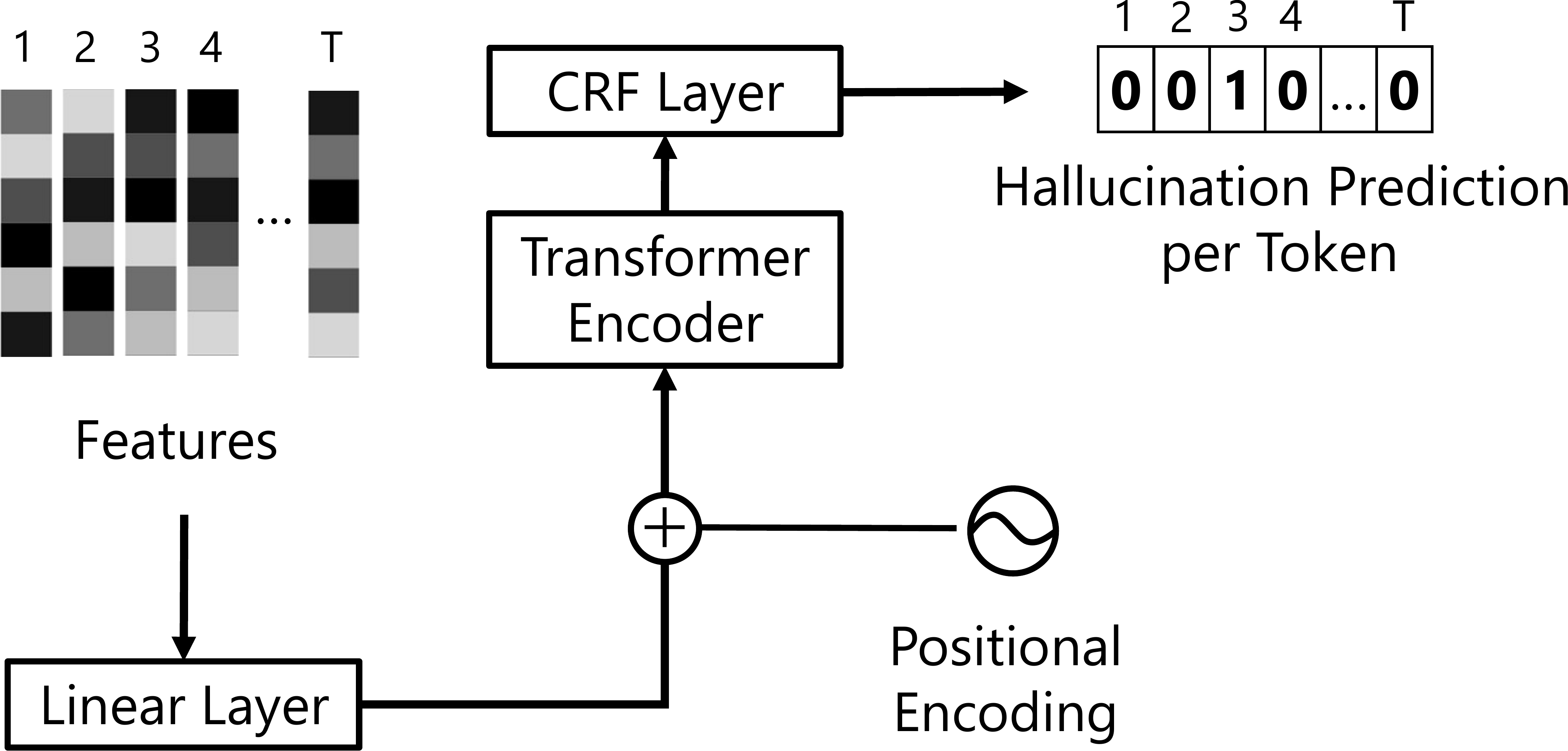}
    \caption{Hallucination Detector}
    \label{fig:detector}
\end{figure}

Our hallucination detector consists of a linear layer, a Transformer encoder layer, and a CRF layer on top, as illustrated in Figure~\ref{fig:detector}. 
To handle \emph{spans}, we employ the CRF layer 
to model dependencies between adjacent tokens, improving the consistency of hallucinated spans compared to independent token-wise classification.\footnote{We empirically confirmed that a linear layer is inferior to CRF in our study.}  
The CRF has been successfully integrated with Transformer-based models 
for structured NLP tasks~\citep{DBLP:journals/corr/abs-1911-04474,wang-etal-2021-hitmi}.  

Feature vectors are first standardised to have $zero$ mean and $1$ standard deviation per feature type. 
After standardisation, the feature vector first goes through a linear layer for transformation, which is primarily employed to adapt to various LLMs that can have different numbers of layers and attention heads. 
Then the transformed vector is input to the transformer layer with positional encoding to incorporate token order information.
Finally, the CRF layer predicts a binary label indicating whether a token is hallucinated (label $1$) or not (label $0$).  
During inference, the Viterbi algorithm determines the most likely label sequences.

\begin{table*}[!t]
\centering
\begin{tabular}{@{}lrrr@{}}
\toprule
\textbf{Dataset}       & \text{QA} & \text{Data2Text} & \text{Summarisation}        \\ \midrule
train         & $4,584$ ($1,421$) ($31.0$\%) & $4,848$ ($3,360$) ($69.3$\%) & $4,308$ ($1,347$) ($31.3$\%)                  \\
valid         & $450$ (~~~$143$) ($31.8$\%) & $450$ (~~~$315$) ($70.0$\%) & $450$ (~~~$135$) ($30.0$\%)                  \\
test            & $900$ (~~~$160$) ($17.8$\%) & $900$ (~~~$579$) ($64.3$\%) & $900$ (~~~$204$) ($22.7$\%)                  \\
Total               & $5,934$ ($1,724$) ($29.1$\%) & $6,198$ ($4,254$) ($68.6$\%) & $5,658$ ($1,686$) ($29.8$\%)                     \\
\bottomrule
\end{tabular}
\caption{Number of samples in the RAGTruth dataset (Numbers in parentheses indicate the raw number of and percentage of sentences containing at least one hallucination span.)}
\label{tab:ragtruth_data}
\end{table*}

\subsection{Attention Weights}
\label{sec:adaptation_kobayashi}
Attention weights have been used to analyse context dependency~\citep{clark-etal-2019-bert,kovaleva-etal-2019-revealing,htut2019attentionheadsberttrack} of Transformer models. 
Recently, \citet{kobayashi-etal-2020-attention} revealed that the norm of the transformed input vector plays a significant role in the attention mechanism. 
They reformulated the computation in the Transformer as: 
\begin{equation}
\bm{y}_i = \sum_{j=1}^T \alpha_{i,j} f(\bm{x}_j)
\label{kobayashi_eq:attention_output}
\end{equation}
where \(\alpha_{i,j}\) is the raw attention weight and \(f(\bm{x}_j)\) is the transformed vector of input \(\bm{x}_j\). 
The transformation function is defined as: 
\begin{equation}
f(\bm{x}) = \left(\bm{x}\bm{W}^V + \bm{b}^V\right)\bm{W}^O,
\label{kobayashi_eq:transformed_value}
\end{equation}
where  \(\bm{W}^V \in \mathbb{R}^{d_{\text{in}} \times d_v}\) and \(\bm{b}^V \in \mathbb{R}^{d_v}\) are the parameters for value transformations and  
\(\bm{W}^O \in \mathbb{R}^{d_v \times d_{\text{out}}}\) is the output matrix multiplication.
\citet{kobayashi-etal-2020-attention} found that frequently occurring tokens often receive high attention weights but have small vector norms, reducing their actual contribution to the output.  
This suggests that attention mechanisms adjust token influence, prioritising informative tokens over frequent but less meaningful ones.  

This study compares the effectiveness of raw and the transformed attention weights of \citet{kobayashi-etal-2020-attention}. 
Specifically, we employ the adjusted attention matrix \(\bm{A}_{\text{norm}}\) defined as:
\begin{equation}
\label{eq:anorm}
\bm{A}_{\text{norm}} = \bm{A} \cdot \mathrm{diag}(\|f(\bm{x})\|),
\end{equation}
where \(\bm{A}\) is the raw attention weight matrix, and \(\mathrm{diag}(\|f(\bm{x})\|)\) represents a diagonal matrix containing the transformed vector norms.


\begin{table}[t]
\centering
\begin{tabular}{@{}ll@{}}
\toprule
\textbf{Hyperparameter}       & \textbf{Search Range}         \\ \midrule
Learning rate            & $1\text{e-}5 \sim 1\text{e-}3$
                  \\
Number of layers               & $[2, 4, 6, 8, 10, 12, 14, 16]$
    \\
Number of heads               & $[4, 8, 16, 32]$
    \\
Dropout rate         & $0.1 \sim 0.5$                     \\
Weight decay            & $1\text{e-}6 \sim 1\text{e-}2$
                  \\
Model dimension            & $[256, 512, 1024]$
                  \\
\midrule
\textbf{Parameter}       & \textbf{Setting}         \\ \midrule
Optimizer                & AdamW                   \\
Batch size  & $64$ (Summrization: $32$)    \\
Maximum epochs & $150$  \\
\bottomrule
\end{tabular}
\caption{Search ranges of Transformer hyperparameters (upper) and training settings (bottom)}
\label{tab:training-encoder_params}
\end{table}

\begin{table*}[t]
\centering
\resizebox{\textwidth}{!}{%
    \begin{tabular}{l|l|ccc|ccc|ccc}
    \toprule
    \multirow{2}{*}{Methods} & \multirow{2}{*}{LLM} 
     & \multicolumn{3}{c}{\text{QA}} & \multicolumn{3}{c}{\text{Data2Text}} & \multicolumn{3}{c}{\text{Summarisation}} \\
     \cmidrule(lr){3-5} \cmidrule(lr){6-8} \cmidrule(lr){9-11} 
     & & Prec & Rec & F1 & Prec & Rec & F1 & Prec & Rec & F1 \\
     \midrule
     Ours$_\textrm{raw}$ & \multirow{5}{*}{Llama} & $47.7$ & $\mathbf{68.7}$ & $56.3$ & $\mathbf{55.6}$ & $55.0$ & $\mathbf{55.3}$ & $51.1$ & $36.7$ & $42.7$ \\
     Ours$_\textrm{norm}$ &  & $57.4$ & $54.0$ & $55.6$ & $53.4$ & $\mathbf{57.1}$ & $55.2$ & $51.0$ & $\mathbf{39.5}$ & $\mathbf{44.5}$ \\
     Fine-tuning &  & $\mathbf{62.8}$ & $56.9$ & $\mathbf{59.7}$ & $55.4$ & $46.2$ & $50.4$ & $\mathbf{52.0}$ & $34.6$ & $41.6$ \\
     Lookback Lens &   & $53.5$ & $7.6$ & $13.2$ & $0.0$ & $0.0$ & $0.0$ & $0.0$ & $0.0$ & $0.0$ \\
    \bottomrule
    \end{tabular}
    }
\caption{Hallucinated span detection results on Llama-3-8B-Instruct. 
The proposed method is denoted as ``Ours'' with variations of raw attention (``raw'') or the transformed attention (``norm''). It outperformed the baselines on tasks with longer input contexts, i.e., Data2Text and Summarisation.
}
\label{tab:5-detection-result}
\end{table*}

\section{Evaluation}
\label{chap:ragtruth_detect}
We evaluate the effectiveness of the proposed method for hallucinated span detection.

\subsection{Dataset}
As the dataset providing hallucination \emph{span} annotation, we employ RAGTruth~\citep{niu-etal-2024-ragtruth}\footnote{\url{https://github.com/ParticleMedia/RAGTruth}}, a benchmark dataset that annotates responses generated by LLMs (GPT-3.5-turbo-0613, GPT-4-0613, Llama-2-7B-chat, Llama-2-13B-chat, Llama-2-70B-chat, and Mistral-7B-Instruct).  
It covers three scenarios of using LLMs in practice, i.e., question answering (QA), data-to-text generation (Data2Text), and news summarisation (Summarisation).  
RAGTruth provides $18,000$ annotated responses, where hallucinated spans in each response are tagged at the character level.  
The number of samples is shown in Table \ref{tab:ragtruth_data}. 
As there is no official validation split in RAGTruth, we randomly sampled $450$ instances ($75$ IDs) from the training set for validation.  

\subsection{Evaluation Metric}
The hallucination labels in RAGTruth are provided at the character span level. 
For example, a hallucination might be annotated with ``start: $219$, end: $229$.'' 
We convert these labels into the token level for intuitive interpretation of evaluation results. 
We employed the same tokeniser of LLM to extract attention matrices.

We compute the token-level precision (Prec) and recall (Rec). 
Given a set of gold-standard hallucination tokens $\mathcal{Y}=\{y_0, y_1, \cdots, y_N\}$ and predicted hallucination tokens $\hat{\mathcal{Y}}=\{\hat{y}_0, \hat{y}_1, \cdots, \hat{y}_M\}$,
\begin{equation}
\text{precision} = \frac{|\hat{\mathcal{Y}} \cap \mathcal{Y}|}{|\hat{\mathcal{Y}}|}, 
\text{recall} = \frac{|\hat{\mathcal{Y}} \cap \mathcal{Y}|}{|\mathcal{Y}|}. 
\end{equation} 
Matching of the gold-standard and predicted tokens is computed in the context of output texts.
The primary evaluation metric is the F1 score of token-level hallucination predictions, which is the harmonic mean of precision and recall. 
Following the RAGTruth evaluation scheme, we used the micro-average of precision, recall, and F1.


\subsection{Implementation}

The proposed method consists of the linear layer, the Transformer encoder layer, and the CRF layer. 
The settings of the Transformer layer, i.e., the numbers of layers and attention heads, the dimensions, and the dropout rate, were tuned together with other hyperparameters of learning rate and weight decay using the Data2Text task, as it provides the largest samples.
We apply the same hyperparameters for other tasks. 
In this study, we used the Optuna library\footnote{\url{https://optuna.org/}} to perform hyperparameter search in the ranges shown in the upper rows of Table~\ref{tab:training-encoder_params}. 
The setting of the model with the highest F1 score was selected for formal evaluation.  

Table~\ref{tab:training-encoder_params} bottom shows training settings: we used AdamW~\citep{loshchilov2019decoupled} optimizer with the batch size of $64$ ($32$ for Summarisation). 
We employed early stopping on training: training was terminated if the F1 score on the validation set did not improve for $10$ consecutive epochs. 
The maximum training epoch was set to $150$.

As the LLM to obtain attention matrices, we employed the recent smaller yet strong models of Llama-3-8B-Instruct~\citep{touvron2023llamaopenefficientfoundation,grattafiori2024llama3herdmodels}
and Qwen2.5-7B-Instruct~\citep{qwen2025qwen25technicalreport} 
(see Appendix~\ref{app:proposed_model} for details). 
We adapted the template by~\citet{niu-etal-2024-ragtruth} for promoting. 
Notice that these LLMs are different from the ones used to create the RAGTruth dataset, which simulates the scenario where we cannot access the LLMs generated outputs for hallucinated span detection.  

\subsection{Baselines}
We compared the proposed method to two baselines employing the same LLMs as our method.

\paragraph{Fine-tuned LLMs}
Although straightforward, fine-tuned LLMs serve as a strong baseline~\citep{niu-etal-2024-ragtruth}. 
We fine-tuned the LLMs using the prompt of~\citet{niu-etal-2024-ragtruth} with instructions to predict hallucinated spans.  
More details are provided in Appendix~\ref{app:fine-tuning}.  

\paragraph{Lookback Lens}  
We employed Lookback Lens~\citep{chuang-etal-2024-lookback}, which also utilises the attention matrix for hallucination detection. 
It computes the ``Lookback'' ratio; the ratio of attention weights on the input context versus newly generated tokens. 
The Lookback feature is input to a logistic regression model to predict the probability of a token being hallucinated.\footnote{Lookback Lens can also conduct span-level prediction by segmenting texts using a sliding window. For direct comparison to our method, we used the token-level variant (i.e., window size is one).} 
We regarded tokens for which the predicted probabilities are equal to or larger than $0.5$ as hallucination, following the traits of the logistic regression classifier. 
We used the author's implementation\footnote{\url{https://github.com/voidism/Lookback-Lens}} for the Lookback Lens model training.

\begin{table}[t]
\centering
\small 
\begin{tabular}{l|rr|rr|rr}
    \toprule
    \multirow{2}{*}{} 
     & \multicolumn{2}{c|}{\text{QA}} & \multicolumn{2}{c|}{\text{Data2Text}} & \multicolumn{2}{c}{\text{Summ.}} \\
     \cmidrule(lr){2-3} \cmidrule(lr){4-5} \cmidrule(lr){6-7} 
     & In & Out & In & Out & In & Out \\
     \midrule
     Mean & $400$ & $140$ & $788$ & $199$ & $723$ & $136$ \\
     Max & $646$ & $437$ & $1,499$ & $406$ & $2,063$ & $412$ \\
     Min & $244$ & $9$ & $517$ & $69$ & $225$ & $16$ \\
    \bottomrule
\end{tabular}
\caption{Numbers of tokens of context (`In') and output (`Out') (measured using Llama-3-8B-Instruct tokeniser).}
\label{tab:ragtruth_token}
\end{table}

\begin{table*}[t]
\centering
\small 
\begin{adjustbox}{max width=\linewidth} 
\begin{tabular}{p{0.9\linewidth}}
\toprule
\textbf{Source text:} 
\texttt{[$\ldots$] From the giant sequoias of Yosemite to the geysers of Yellowstone, the United States' national parks were made for you and me. And for Saturday and Sunday, they're also free. Though most of the National Park Service's 407 sites are free year-round, the 128 parks that charge a fee -- like Yellowstone and Yosemite -- will be free those two days. It's all part of National Park Week, happening April 18 through April 26, and it's hosted by the National Park Service and the National Park Foundation. [$\ldots$] } \\\hdashline
\textbf{Output summary:} \texttt{National Park Service offers free admission to 128 parks, including Yellowstone and Yosemite, \textred{on April 18-19 and 25-26}, as part of National Park Week.} \\
\hline
\textbf{Ground Truth:} \texttt{on April 18-19 and 25-26} \\
\textbf{Ours$_\textrm{raw}$:} \texttt{April 18-19 and 25-26}  \\
\textbf{Fine-tuning:} \texttt{--} (Detection failed) \\
\bottomrule
\end{tabular}
\end{adjustbox}
\caption{Hallucination detection example (Summarisation)}
\label{tab:rag_output_sum}
\end{table*}

\begin{table*}[t]
\centering
\resizebox{\linewidth}{!}{%
    \begin{tabular}{l|cccccccccccc}
    \toprule
    \multirow{2}{*}{Methods}
     & \multicolumn{4}{c}{\text{QA}} & \multicolumn{4}{c}{\text{Data2Text}} & \multicolumn{4}{c}{\text{Summarisation}} \\
     \cmidrule(lr){2-5} \cmidrule(lr){6-9} \cmidrule(lr){10-13} 
     & 0–2 & 2–4 & 4–6 & 6–8
     & 0–2 & 2–4 & 4–6 & 6–8 
     & 0–2 & 2–4 & 4–6 & 6–8 \\
     \midrule
     Ours$_\textrm{raw}$ & $27.7$ & $-$ & $48.6$ & $59.4$ & $\mathbf{33.0}$ & $-$ & $\mathbf{52.6}$ & $\mathbf{63.3}$ & $0.0$ & $\mathbf{42.3}$ & $28.5$ & $54.4$\\
     Ours$_\textrm{norm}$ & $25.1$ & $-$ & $41.1$ & $61.0$ & $\mathbf{33.0}$ & $-$ & $51.2$ & $61.9$ & $0.0$ & $41.9$ & $30.5$ & $\mathbf{59.0}$\\
     Fine-tuning & $\mathbf{38.4}$ & $-$ & $\mathbf{52.7}$ & $\mathbf{62.3}$ & $23.8$ & $-$ & $45.8$ & $57.9$ & $0.0$ & $41.0$ & $\mathbf{31.4}$ & $56.4$\\
    \bottomrule
    \end{tabular}
    }
\caption{Token-level F1 scores of hallucinated span detection per different hallucination ratios (Llama-3-8B-Instruct). 
``$-$'' indicates there was no sample falling in the corresponding bin.}
\label{tab:5-bin-result}
\end{table*}

\subsection{Experimental Results} 
The experimental results on Llama-3-8B-Instruct are shown in Table~\ref{tab:5-detection-result}. 
The proposed method is denoted as ``Ours'' with variations of using raw attention weights (denoted as ``raw'') and the transformed attention weights (denoted as ``norm''). 

The proposed method outperformed both the fine-tuning and Lookback Lens for hallucinated span detection in Data2Text and summarisation, achieving the highest token-level F1 scores. 
On QA, the proposed method tends to have higher recall yet lower precision, i.e., it tends to overly detect hallucinations. 
A possible factor is shorter lengths of input context. 
Table~\ref{tab:ragtruth_token} shows the numbers of tokens in context and output texts. 
QA has significantly shorter contexts on average compared to Data2Text and summarisation, while the output lengths are similar. 
This result may imply that the proposed method better handles tasks where consistency with long context is important, like summarisation. 
We conduct further analysis in Sections~\ref{sec:effect_token_rate} and~\ref{sec:performance_hallucination_type}.

For attention weights, the effectiveness of the raw and transformed attention weights depends on tasks. 
The raw attention weights performed higher in QA, while the transformed weights outperformed the raw attention in summarisation, and they are comparable on Data2Text.

Lookback Lens consistently exhibited the lowest F1 scores.\footnote{This looks largely different from the original paper. We remark that in addition to the experimental dataset difference, the original paper reported AUROC.} 
Our inspection confirmed that Lookback Lens overfitted the majority class, i.e., no hallucination. 
Hallucinated spans are much more infrequent compared to the no-hallucination tokens. 
This implies that making a binary decision based on the predicted hallucination probability is non-trivial.
Furthermore, Lookback Lens seems to have struggled to handle longer input contexts, i.e., Data2Text and summarisation tasks, in contrast to the proposed method. 
This may be because the Lookback Lens strongly depends on attention weights for the input context. 
We evaluated the combination of features of Lookback Lens and ours to see if they are complementary. 
As a result, no improvement was observed; possibly because our ``Outgoing Attention Entropy'' feature also takes the input context into account. 
Table \ref{tab:rag_output_sum} presents an example of hallucination detection on summarisation. 
In the output text, the red-coloured span indicates the hallucination. 
While the Fine-tuning failed to detect the hallucination, the proposed method successfully identified the span very close to the ground truth (only missing a preposition). 
Further examples are in Appendix~\ref{app:hal_examples}.

\begin{table*}[t]
\centering
\resizebox{\textwidth}{!}{%
    \begin{tabular}{l|l|ccc|ccc|ccc}
    \toprule
    \multirow{2}{*}{Methods} & \multirow{2}{*}{LLM} 
     & \multicolumn{3}{c}{\text{QA}} & \multicolumn{3}{c}{\text{Data2Text}} & \multicolumn{3}{c}{\text{Summarisation}} \\
     \cmidrule(lr){3-5} \cmidrule(lr){6-8} \cmidrule(lr){9-11} 
     & & Prec & Rec & F1 & Prec & Rec & F1 & Prec & Rec & F1 \\
     \midrule
     Ours$_\textrm{raw}$ & \multirow{5}{*}{Qwen} & $38.5$ & $\mathbf{73.7}$ & $50.6$ & $53.5$ & $\mathbf{57.1}$ & $55.2$ & $49.6$ & $\mathbf{35.7}$ & $\mathbf{41.5}$ \\
     Ours$_\textrm{norm}$ &   & $39.0$ & $64.7$ & $48.7$ & $55.5$ & $55.3$ & $\mathbf{55.4}$ & $49.3$ & $33.6$ & $39.9$ \\
     Fine-tuning &   & $\mathbf{60.1}$ & $57.1$ & $\mathbf{58.6}$ & $\mathbf{58.9}$ & $51.4$ & $54.9$ & $\mathbf{62.0}$ & $30.0$ & $40.4$ \\
     Lookback Lens &   & $46.6$ & $5.6$ & $9.9$ & $50.0$ & $0.0$ & $0.0$ & $0.0$ & $0.0$ & $0.0$ \\
    \bottomrule
    \end{tabular}
}
\caption{Hallucinated span detection results on Qwen2.5-7B-Instruct}
\label{tab:5-detection-result-qwen}
\end{table*}

\begin{table}[t]
\centering
\begin{adjustbox}{max width=\linewidth}
\begin{tabular}{l|cccc|c}
\hline
\multicolumn{6}{c}{QA (Total Tokens: 124,817)} \\
Methods & SInfo & EInfo & SConf & EConf & All \\
\hline
Ours$_\textrm{raw}$ & $\mathbf{74.1}$ & $\mathbf{74.4}$ & $-$ & $4.0$ & $\mathbf{68.7}$ \\
Ours$_\textrm{norm}$ & $50.6$ & $60.0$ & $-$ & $3.8$ & $54.0$ \\
Fine-tuning & $48.7$ & $63.8$ & $-$ & $\mathbf{7.8}$ & $56.9$ \\
\hdashline
Hal. Tokens & $1,020$ & $4,742$ & $-$ & $501$ & $6,263$ \\
\hline
\multicolumn{6}{c}{Data2Text (Total Tokens: 178,343)} \\
Methods & SInfo & EInfo & SConf & EConf & All \\
\hline
Ours$_\textrm{raw}$ & $29.4$ & $50.5$ & $\mathbf{7.3}$ & $64.7$ & $55.5$ \\
Ours$_\textrm{norm}$ & $\mathbf{37.8}$ & $\mathbf{52.7}$ & $\mathbf{7.3}$ & $\mathbf{64.8}$ & $\mathbf{57.1}$ \\
Fine-tuning & $35.8$ & $51.6$ & $0.0$ & $43.7$ & $46.2$ \\
\hdashline
Hal. Tokens & $595$ & $3,118$ & $41$ & $3,580$ & $7,334$ \\
\hline
\multicolumn{6}{c}{Summarisation (Total Tokens: 121,248)} \\
Methods & SInfo & EInfo & SConf & EConf & All \\
\hline
Ours$_\textrm{raw}$ & $\mathbf{65.2}$ & $46.5$ & $\mathbf{8.5}$ & $16.4$ & $36.7$ \\
Ours$_\textrm{norm}$ & $49.7$ & $\mathbf{51.3}$ & $\mathbf{8.5}$ & $18.5$ & $\mathbf{39.5}$ \\
Fine-tuning & $44.9$ & $43.7$ & $8.1$ & $\mathbf{18.6}$ & $34.6$ \\
\hdashline
Hal. Tokens & $187$ & $2,067$ & $71$ & $1,160$ & $3,485$ \\
\hline
\end{tabular}
\end{adjustbox}
\caption{Recall of hallucinated span detection per hallucination type (Llama-3-8B-Instruct)}
\label{tab:hallucination-detection-results}
\end{table}

\subsection{Effects of Hallucination Ratio}
\label{sec:effect_token_rate}

Intuitively, the ratio of hallucinated tokens in a text affects the performance. 
When the frequency of hallucinations is small, detection should become more challenging. 
Table \ref{tab:5-bin-result} shows the token-level F1 scores on different percentages of hallucinated tokens. 
These results confirm that the intuition holds true. 
Across methods and tasks, higher F1 scores were achieved when hallucinated tokens were more frequent. 

Another interesting observation is that the effect of task type is dominant than the hallucinated token ratio. 
Table \ref{tab:5-bin-result} shows that the superior method is consistent across different frequencies of hallucinated tokens within the same task. 

\subsection{Effects of Hallucination Type}
\label{sec:performance_hallucination_type}
We further analysed the hallucination detection capability of the proposed method for different hallucination types. 
RAGTruth categorises hallucinations into four types: 
Subtle Introduction of Baseless Information (\textbf{SInfo}) and Evident Introduction of Baseless Information (\textbf{EInfo}) indicate whether the output text subtly adds information or explicitly introduces falsehoods.
Subtle Conflict (\textbf{SConf}) and Evident Conflict (\textbf{EConf}) indicate whether the output alters meaning or directly contradicts the input text.  
For more details, see~\citet{niu-etal-2024-ragtruth}.

Table~\ref{tab:hallucination-detection-results} shows detection recalls for different hallucination types.\footnote{Precision (and thus F1) is difficult to compute because it is non-trivial to decide to which category does detected hallucination belong.} 
For Data2Text, the recall of Evident Conflict is significantly higher than SInfo and EInfo. 
This result indicates that the proposed method better captures conflicting information against input context than baseless information introduced by LLMs. 
The trend is the opposite on QA and summarisation, where the proposed method achieved much higher recall on SInfo and EInfo than on SConf and EConf, which implies that baseless information was easier to capture for the proposed method. 
These results indicate that detection difficulties of different hallucination types can vary depending on tasks.


\subsection{Performance on Qwen}
Table~\ref{tab:5-detection-result-qwen} shows the results on Qwen2.5-7B-Instruct.
While the results are consistent with Table~\ref{tab:5-detection-result}, Qwen was consistently inferior to Llama regarding the proposed method, which should be attributed to different implementations of their attention mechanisms. 
Specifically, Llama-3-8B-Instruct has $32$ layers and $32$ attention heads, while Qwen2.5-7B-Instruct has $28$ layers and $28$ heads.
Qwen has fewer numbers of layers and attention heads, and thus its feature dimension is smaller than Llama. 
In addition, the parameters in multi-head attention are more aggressively shared in Qwen. 
These differences may affect the attention features extracted from Qwen. 
More details of the differences between Llama and Qwen are discussed in Appendix~\ref{app:proposed_model}.

\section{Conclusion}
We proposed the hallucinated span detection method using features that assemble attention weights from different views. 
Our experiments confirmed that these features are useful in combination for detecting hallucinated spans, outperforming a previous method that also uses attention weights.  


This study focused on hallucination detection, but our method may also apply to broader abnormal behaviour detection of LLMs. 
As future work, we plan to explore its potential for detecting backdoored LLMs~\citep{he2023mitigating}, which behave normally on regular inputs but produce malicious outputs when triggered. 
Since our approach analyses attention distributions, it may detect anomalous attention patterns caused by the triggers. 


\section*{Limitations}
While we confirmed the effectiveness of the proposed method on two models: Llama-3-8B-Instruct and Qwen2.5-7B-Instruct, there are lots more LLMs. 
The effectiveness of our method when applied to attention mechanisms from other models remains unverified. 
In addition, our experiments are limited to the English language. 
We will explore the applicability of our method to other languages by employing multilingual LLMs. 

Our method requires training data that annotates hallucinated spans, which is costly to create. 
A potential future direction is an exploration of an unsupervised learning approach. 
The success of the current method implies that our features successfully capture irregular attention patterns on hallucination. 
We plan to train our method only on non-hallucinated human-written text. 
We then identify hallucinations as instances in which attention patterns deviate from the learned normal patterns.  


\section*{Acknowledgement}
We sincerely thank Professor Tomoyuki Kajiwara for his insightful comments and valuable discussions that greatly improved this work.
This work was supported by JST K Program Grant Number JPMJKP$24$C$3$, Japan.

\bibliography{custom}

\appendix

\section{Details of Experiment Settings}
\label{app:experiments}

\subsection{Computational Environment}
All the experiments were conducted on NVIDIA RTX A$6000$ ($48$GB memory) GPUs.
For training the Transformer encoder of the proposed method, we used 2 GPUs. 
For fine-tuning the LLM, we used 4 GPUs in parallel.

\subsection{LLM Details}
\label{app:proposed_model}
Llama-3-8B-Instruct has $32$ layers and $32$ attention heads, while Qwen2.5-7B-Instruct has $28$ layers and $28$ heads.  
Both models replace standard Multi-Head Attention (MHA) with Grouped-Query Attention (GQA)~\citep{ainslie-etal-2023-gqa}, but Llama-3 uses more layers and heads than Qwen2.5.  

MHA assigns each query to a single key-value pair, whereas GQA allows multiple queries to share a key-value pair, reducing the number of trainable parameters.  
Llama-3-8B-Instruct processes $32$ queries while reducing the number of keys and values to $8$, so each key-value pair corresponds to $4$ queries.  
In contrast, Qwen2.5-7B-Instruct processes $28$ queries and reduces the number of keys and values to $4$, making each key-value pair correspond to $7$ queries.  

We conjecture these differences were reflected in the different performances of Llama and Qwen in our method.

\subsection{Fine-Tuning}
\label{app:fine-tuning}
\begin{table}[h!]
\centering
\begin{tabular}{@{}ll@{}}
\toprule
\textbf{Parameter}       & \textbf{Value}         \\ \midrule
Fine-tuning method            & full fine-tuning                  \\
Learning rate            & $5$e-$6$                  \\
Batch size               & $1$                     \\
Number of epochs         & $3$                     \\
Optimizer                & AdamW                   \\
Warmup steps            & $10$     \\
\bottomrule
\end{tabular}
\caption{Fine-tuning Parameters}
\label{tab:finetuing-params}
\end{table}

Fine-tuning was conducted using LLaMA-Factory~\citep{zheng-etal-2024-llamafactory}\footnote{\url{https://github.com/hiyouga/LLaMA-Factory}}, a library specialized for fine-tuning LLMs.  
The fine-tuning parameters are shown in Table~\ref{tab:finetuing-params}.  
The fine-tuned model predicts the hallucinated span by predicting character indexes. 
If a hallucination label changes within a single token in predictions, the entire token is considered as being hallucinated.

\subsection{Prompts of RAGTruth}
The prompts used in our experiments are shown in Table~\ref{tab:llama_rag_prompt} and Table~\ref{tab:qwen_rag_prompt}.

\begin{table*}[t]
\centering
\small 
\begin{adjustbox}{max width=\linewidth} 
\begin{tabular}{p{0.9\linewidth}}
\toprule
\multicolumn{1}{c}{\textbf{QA Prompt}} \\ 
\midrule
\textbf{Original text (including tokens):} \\
\texttt{<|begin\_of\_text|><|start\_header\_id|>system<|end\_header\_id|>} \\
\texttt{You are an excellent system, generating output according to the instructions.} \\
\texttt{<|eot\_id|><|start\_header\_id|>user<|end\_header\_id|>} \\
\texttt{Briefly answer the following question:} \\
\texttt{\{question\}} \\
\texttt{Bear in mind that your response should be strictly based on the following three passages:} \\
\texttt{\{passages\}} \\
\texttt{In case the passages do not contain the necessary information to answer the question, please reply with:} \\
\texttt{"Unable to answer based on given passages."} \\
\texttt{output:} \\
\texttt{<|eot\_id|><|start\_header\_id|>assistant<|end\_header\_id|>} \\
\texttt{\{answer\} <|eot\_id|>} \\
\midrule
\multicolumn{1}{c}{\textbf{Data2Text Prompt}} \\ 
\midrule
\textbf{Original text (including tokens):} \\
\texttt{<|begin\_of\_text|><|start\_header\_id|>system<|end\_header\_id|>} \\
\texttt{You are an excellent system, generating output according to the instructions.} \\
\texttt{<|eot\_id|><|start\_header\_id|>user<|end\_header\_id|>} \\
\texttt{Instruction:} \\
\texttt{Write an objective overview about the following local business based only on the provided structured data in the JSON format.} \\
\texttt{You should include details and cover the information mentioned in the customers’ review.} \\
\texttt{The overview should be 100 - 200 words. Don’t make up information.} \\
\texttt{Structured data:} \\
\texttt{\{json\_data\}} \\
\texttt{Overview:} \\
\texttt{<|eot\_id|><|start\_header\_id|>assistant<|end\_header\_id|>} \\
\texttt{\{Converted text\} <|eot\_id|>} \\
\midrule
\multicolumn{1}{c}{\textbf{Summarisation Prompt}} \\ 
\midrule
\textbf{Original text (including tokens):} \\
\texttt{<|begin\_of\_text|><|start\_header\_id|>system<|end\_header\_id|>} \\
\texttt{You are an excellent system, generating output according to the instructions.} \\
\texttt{<|eot\_id|><|start\_header\_id|>user<|end\_header\_id|>} \\
\texttt{Summarize the following news within \{word count of the summary\} words:} \\
\texttt{\{text to summarize\}} \\
\texttt{output:} \\
\texttt{<|eot\_id|><|start\_header\_id|>assistant<|end\_header\_id|>} \\
\texttt{\{summary\} <|eot\_id|>} \\
\bottomrule
\end{tabular}
\end{adjustbox}
\caption{Prompts for RAGTruth (Using Llama-3-8B-Instruct)}
\label{tab:llama_rag_prompt}
\end{table*}

\begin{table*}[t]
\centering
\small 
\begin{adjustbox}{max width=\linewidth} 
\begin{tabular}{p{0.9\linewidth}}
\toprule
\multicolumn{1}{c}{\textbf{QA Prompt}} \\ 
\midrule
\textbf{Original text (including tokens):} \\
\texttt{<\textbar im\_start\textbar>system} \\
\texttt{You are an excellent system, generating output according to the instructions.<\textbar im\_end\textbar>} \\
\texttt{<\textbar im\_start\textbar>user} \\
\texttt{Briefly answer the following question:} \\
\texttt{\{question\}} \\
\texttt{Bear in mind that your response should be strictly based on the following three passages:} \\
\texttt{\{passages\}} \\
\texttt{In case the passages do not contain the necessary information to answer the question, please reply with:} \\
\texttt{"Unable to answer based on given passages."} \\
\texttt{output:<\textbar im\_end\textbar>} \\
\texttt{<\textbar im\_start\textbar>assistant} \\
\texttt{\{answer\}<\textbar im\_end\textbar>} \\
\midrule
\multicolumn{1}{c}{\textbf{Data2Text Prompt}} \\ 
\midrule
\textbf{Original text (including tokens):} \\
\texttt{<\textbar im\_start\textbar>system} \\
\texttt{You are an excellent system, generating output according to the instructions.<\textbar im\_end\textbar>} \\
\texttt{<\textbar im\_start\textbar>user} \\
\texttt{Instruction:} \\
\texttt{Write an objective overview about the following local business based only on the provided structured data in the JSON format.} \\
\texttt{You should include details and cover the information mentioned in the customers’ review.} \\
\texttt{The overview should be 100 - 200 words. Don’t make up information.} \\
\texttt{Structured data:} \\
\texttt{\{json\_data\}} \\
\texttt{Overview:<\textbar im\_end\textbar>} \\
\texttt{<\textbar im\_start\textbar>assistant} \\
\texttt{\{Converted text\}<\textbar im\_end\textbar>} \\
\midrule
\multicolumn{1}{c}{\textbf{Summarisation Prompt}} \\ 
\midrule
\textbf{Original text (including tokens):} \\
\texttt{<\textbar im\_start\textbar>system} \\
\texttt{You are an excellent system, generating output according to the instructions.<\textbar im\_end\textbar>} \\
\texttt{<\textbar im\_start\textbar>user} \\
\texttt{Summarize the following news within \{word count of the summary\} words:} \\
\texttt{\{text to summarize\}} \\
\texttt{output:<\textbar im\_end\textbar>} \\
\texttt{<\textbar im\_start\textbar>assistant} \\
\texttt{\{summary\}<\textbar im\_end\textbar>} \\
\bottomrule
\end{tabular}
\end{adjustbox}
\caption{Prompts for RAGTruth (Using Qwen2.5-7B-Instruct)}
\label{tab:qwen_rag_prompt}
\end{table*}


\section{Hallucination Detection Examples}
\label{app:hal_examples}

Table \ref{tab:rag_output_qa} presents hallucination detection results in the QA task. 
The Fine-tuning baseline incorrectly judged the non-hallucinated span as hallucinated and largely overlooked the truly hallucinated span. 
In contrast, the proposed method mostly correctly identified the hallucinated span. 

Table \ref{tab:rag_output_sum_miss} presents hallucination detection results in the summarisation task where the proposed method failed. 
In the first example, the proposed method overlooked the hallucinated span. 
In the second example, the proposed method mistook the non-hallucinated span as hallucinated.

\begin{table*}[t]
\centering
\small 
\begin{adjustbox}{max width=\linewidth} 
\begin{tabular}{p{0.9\linewidth}}
\toprule
\textbf{Key Part of the Question} \\
\texttt{Briefly answer the following question: differences between elements, compounds, and mixtures. Bear in mind that your response should be strictly based on the following three passages: ...
} (Following this, three passages appear, but there is no information on the bonding of elements or compounds. The content primarily focuses on mixtures.) \\
\hdashline
\textbf{Ground Truth Label} \\
\texttt{...
3. Separation: Elements can be separated easily by physical means, while separating compounds requires more complex methods like chemical reactions. It is easy to physically separate the components of a mixture. \textred{4. Bonding: Elements are held together by covalent bonds, while compounds can have various types of bonds (covalent, ionic, metallic, etc.). Mixtures do not have any bonds between their components.}}  \\
\hdashline
\textbf{Ours$_\textrm{raw}$} \\
\texttt{...
3. Separation: Elements can be separated easily by physical means, while separating compounds requires more complex methods like chemical reactions. It is easy to physically separate the components of a mixture. 4. \textred{Bonding: Elements are held together by covalent bonds, while compounds can have various types of bonds (covalent, ionic, metallic, etc.). Mixtures do not have any bonds between their components.}}  \\
\hdashline
\textbf{Fine-tuning} \\
\texttt{...
\textred{3. Separation: Elements can be separated easily by physical means, while separating compounds requires more complex methods like chemical reactions.} It is easy to physically separate the components of a mixture. 4. Bonding: Elements are held together by covalent bonds, while compounds can have various types of bonds (covalent, ionic, metallic, etc.). \textred{Mixtures do not have any bonds between their components.}} (Incorrect detection in point 3 + incomplete detection in point 4. The discussion on separation in point 3 is fully described in the original text.)\\
\bottomrule
\end{tabular}
\end{adjustbox}
\caption{Hallucination detection example (QA)}
\label{tab:rag_output_qa}
\end{table*}

\begin{table*}[t]
\centering
\small 
\begin{adjustbox}{max width=\linewidth} 
\begin{tabular}{p{0.9\linewidth}}
\toprule
\multicolumn{1}{c}{\textbf{Example 1}} \\ 
\midrule
\textbf{Key Part of the Target Sentence for Summarisation} \\
\texttt{... Doug Ducey signed legislation to allow Arizonans to get any lab test without a doctor's order. Freedom of information -- always sounds like a good thing. ...} (The target sentence for summarisation contains no mention of Doug Ducey being the governor of Texas. In fact, he was a former governor of Arizona, making this incorrect.)\\
\hdashline
\textbf{Ground Truth Label} \\
\texttt{The article discusses the increasing trend of individuals getting tested for various medical conditions without a prescription. \textred{Texas Governor Doug Ducey} has signed legislation allowing Arizonans to get any lab test they desire without consulting a doctor first. ...}  \\
\hdashline
\textbf{Ours$_\textrm{raw}$} \\
\texttt{The article discusses the increasing trend of individuals getting tested for various medical conditions without a prescription. Texas Governor Doug Ducey has signed legislation allowing Arizonans to get any lab test they desire without consulting a doctor first. ...} (Detection failed) \\
\hdashline
\textbf{Fine-tuning} \\
\texttt{The article discusses the increasing trend of individuals getting tested for various medical conditions without a prescription. \textred{Texas Governor Doug Ducey} has signed legislation allowing Arizonans to get any lab test they desire without consulting a doctor first. ...} \\
\midrule
\multicolumn{1}{c}{\textbf{Example 2}} \\ 
\midrule
\textbf{Key Part of the Target Sentence for summarisation} \\
\texttt{... Still, the average monthly benefit for retired workers rising by \$59 to \$1,907 will undoubtedly help retirees with lower and middle incomes to better cope with inflation. ...} (\$$1907$-\$$59$=\$$1848$ increase)\\
\hdashline
\textbf{Ground Truth Label} \\
\texttt{... Retired workers can expect an average monthly benefit of \$1,907, up from \$1,848. ...}  \\
\hdashline
\textbf{Ours$_\textrm{raw}$} \\
\texttt{... Retired workers can expect an average \textred{monthly benefit of \$1,907, up from \$1,848}. ...} (False detection) \\
\hdashline
\textbf{Fine-tuning} \\
\texttt{... Retired workers can expect an average monthly benefit of \$1,907, up from \$1,848. ...} \\
\bottomrule
\end{tabular}
\end{adjustbox}
\caption{Hallucination detection example (Summarisation)}
\label{tab:rag_output_sum_miss}
\end{table*}

\end{document}